\documentclass[]{article}
\usepackage{lmodern}
\usepackage{svg}
\usepackage{amssymb,amsmath}
\usepackage{ifxetex,ifluatex}
\usepackage{fixltx2e} 
\ifnum 0\ifxetex 1\fi\ifluatex 1\fi=0 
  \usepackage[T1]{fontenc}
  \usepackage[utf8]{inputenc}
\else 
  \ifxetex
    \usepackage{mathspec}
  \else
    \usepackage{fontspec}
  \fi
  \defaultfontfeatures{Ligatures=TeX,Scale=MatchLowercase}
\fi
\IfFileExists{upquote.sty}{\usepackage{upquote}}{}
\IfFileExists{microtype.sty}{%
\usepackage{microtype}
\UseMicrotypeSet[protrusion]{basicmath} 
}{}
\usepackage[margin=2cm]{geometry}
\usepackage[unicode=true]{hyperref}
\hypersetup{
            pdftitle={Data Science and Digital Systems: The 3Ds of Machine Learning Systems Design},
            pdfborder={0 0 0},
            breaklinks=true}
\usepackage{natbib}
\IfFileExists{parskip.sty}{%
\usepackage{parskip}
}{
\setlength{\parindent}{0pt}
\setlength{\parskip}{6pt plus 2pt minus 1pt}
}
\makeatletter

\def\fps@figure{htbp}
\makeatother
\usepackage{graphicx}

\usepackage{amssymb} 
\usepackage{enumitem}
\usepackage{xcolor}
\usepackage{lipsum}
\usepackage{titlesec}

\providecommand{\tightlist}{%
  \setlength{\itemsep}{0pt}\setlength{\parskip}{0pt}}

\begin{document}

\title{Data Science and Digital Systems: The 3Ds of Machine Learning Systems
Design}

\author{Neil D. Lawrence}

\maketitle

\abstract{Machine learning solutions, in particular those based on deep learning
methods, form an underpinning of the current revolution in ``artificial
intelligence'' that has dominated popular press headlines and is having
a significant influence on the wider tech agenda. Here we give an
overview of the 3Ds of ML systems design: Data, Design and Deployment.
By considering the 3Ds we can move towards \emph{data first} design.}

\newcommand{\Amatrix}{\mathbf{A}}
\newcommand{\KL}[2]{\text{KL}\left( #1\,\|\,#2 \right)}
\newcommand{\Kaast}{\kernelMatrix_{\mathbf{ \ast}\mathbf{ \ast}}}
\newcommand{\Kastu}{\kernelMatrix_{\mathbf{ \ast} \inducingVector}}
\newcommand{\Kff}{\kernelMatrix_{\mappingFunctionVector \mappingFunctionVector}}
\newcommand{\Kfu}{\kernelMatrix_{\mappingFunctionVector \inducingVector}}
\newcommand{\Kuast}{\kernelMatrix_{\inducingVector \bf\ast}}
\newcommand{\Kuf}{\kernelMatrix_{\inducingVector \mappingFunctionVector}}
\newcommand{\Kuu}{\kernelMatrix_{\inducingVector \inducingVector}}
\newcommand{\Kuui}{\Kuu^{-1}}
\newcommand{\Qaast}{\mathbf{Q}_{\bf \ast \ast}}
\newcommand{\Qastf}{\mathbf{Q}_{\ast \mappingFunction}}
\newcommand{\Qfast}{\mathbf{Q}_{\mappingFunctionVector \bf \ast}}
\newcommand{\Qff}{\mathbf{Q}_{\mappingFunctionVector \mappingFunctionVector}}
\newcommand{\aMatrix}{\mathbf{A}}
\newcommand{\aScalar}{a}
\newcommand{\aVector}{\mathbf{a}}
\newcommand{\acceleration}{a}
\newcommand{\bMatrix}{\mathbf{B}}
\newcommand{\bScalar}{b}
\newcommand{\bVector}{\mathbf{b}}
\newcommand{\basisFunc}{\phi}
\newcommand{\basisFuncVector}{\boldsymbol{ \basisFunc}}
\newcommand{\basisFunction}{\phi}
\newcommand{\basisLocation}{\mu}
\newcommand{\basisMatrix}{\boldsymbol{ \Phi}}
\newcommand{\basisScalar}{\basisFunction}
\newcommand{\basisVector}{\boldsymbol{ \basisFunction}}
\newcommand{\activationFunction}{\phi}
\newcommand{\activationMatrix}{\boldsymbol{ \Phi}}
\newcommand{\activationScalar}{\basisFunction}
\newcommand{\activationVector}{\boldsymbol{ \basisFunction}}
\newcommand{\bigO}{\mathcal{O}}
\newcommand{\binomProb}{\pi}
\newcommand{\cMatrix}{\mathbf{C}}
\newcommand{\cbasisMatrix}{\hat{\boldsymbol{ \Phi}}}
\newcommand{\cdataMatrix}{\hat{\dataMatrix}}
\newcommand{\cdataScalar}{\hat{\dataScalar}}
\newcommand{\cdataVector}{\hat{\dataVector}}
\newcommand{\centeredKernelMatrix}{\mathbf{ \MakeUppercase{\centeredKernelScalar}}}
\newcommand{\centeredKernelScalar}{b}
\newcommand{\centeredKernelVector}{\centeredKernelScalar}
\newcommand{\centeringMatrix}{\mathbf{H}}
\newcommand{\chiSquaredDist}[2]{\chi_{#1}^{2}\left(#2\right)}
\newcommand{\chiSquaredSamp}[1]{\chi_{#1}^{2}}
\newcommand{\conditionalCovariance}{\boldsymbol{ \Sigma}}
\newcommand{\coregionalizationMatrix}{\mathbf{B}}
\newcommand{\coregionalizationScalar}{b}
\newcommand{\coregionalizationVector}{\mathbf{ \coregionalizationScalar}}
\newcommand{\covDist}[2]{\text{cov}_{#2}\left(#1\right)}
\newcommand{\covSamp}[1]{\text{cov}\left(#1\right)}
\newcommand{\covarianceScalar}{c}
\newcommand{\covarianceVector}{\mathbf{ \covarianceScalar}}
\newcommand{\covarianceMatrix}{\mathbf{C}}
\newcommand{\covarianceMatrixTwo}{\boldsymbol{ \Sigma}}
\newcommand{\croupierScalar}{s}
\newcommand{\croupierVector}{\mathbf{ \croupierScalar}}
\newcommand{\croupierMatrix}{\mathbf{ \MakeUppercase{\croupierScalar}}}
\newcommand{\dataDim}{p}
\newcommand{\dataIndex}{i}
\newcommand{\dataIndexTwo}{j}
\newcommand{\dataMatrix}{\mathbf{Y}}
\newcommand{\dataScalar}{y}
\newcommand{\dataSet}{\mathcal{D}}
\newcommand{\dataStd}{\sigma}
\newcommand{\dataVector}{\mathbf{ \dataScalar}}
\newcommand{\decayRate}{d}
\newcommand{\degreeMatrix}{\mathbf{ \MakeUppercase{\degreeScalar}}}
\newcommand{\degreeScalar}{d}
\newcommand{\degreeVector}{\mathbf{ \degreeScalar}}
\newcommand{\diag}[1]{\text{diag}\left(#1\right)}
\newcommand{\diagonalMatrix}{\mathbf{D}}
\newcommand{\diff}[2]{\frac{\text{d}#1}{\text{d}#2}}
\newcommand{\diffTwo}[2]{\frac{\text{d}^2#1}{\text{d}#2^2}}
\newcommand{\displacement}{x}
\newcommand{\displacementVector}{\textbf{\displacement}}
\newcommand{\distanceMatrix}{\mathbf{ \MakeUppercase{\distanceScalar}}}
\newcommand{\distanceScalar}{d}
\newcommand{\distanceVector}{\mathbf{ \distanceScalar}}
\newcommand{\eigenvaltwo}{\ell}
\newcommand{\eigenvaltwoMatrix}{\mathbf{L}}
\newcommand{\eigenvaltwoVector}{\mathbf{l}}
\newcommand{\eigenvalue}{\lambda}
\newcommand{\eigenvalueMatrix}{\boldsymbol{ \Lambda}}
\newcommand{\eigenvalueVector}{\boldsymbol{ \lambda}}
\newcommand{\eigenvector}{\mathbf{ \eigenvectorScalar}}
\newcommand{\eigenvectorMatrix}{\mathbf{U}}
\newcommand{\eigenvectorScalar}{u}
\newcommand{\eigenvectwo}{\mathbf{v}}
\newcommand{\eigenvectwoMatrix}{\mathbf{V}}
\newcommand{\eigenvectwoScalar}{v}
\newcommand{\entropy}[1]{\mathcal{H}\left(#1\right)}
\newcommand{\errorFunction}{E}
\newcommand{\expDist}[2]{\left<#1\right>_{#2}}
\newcommand{\expSamp}[1]{\left<#1\right>}
\newcommand{\expectation}[1]{\left\langle #1 \right\rangle }
\newcommand{\expectationDist}[2]{\left\langle #1 \right\rangle _{#2}}
\newcommand{\expectedDistanceMatrix}{\mathcal{D}}
\newcommand{\eye}{\mathbf{I}}
\newcommand{\fantasyDim}{r}
\newcommand{\fantasyMatrix}{\mathbf{ \MakeUppercase{\fantasyScalar}}}
\newcommand{\fantasyScalar}{z}
\newcommand{\fantasyVector}{\mathbf{ \fantasyScalar}}
\newcommand{\featureStd}{\varsigma}
\newcommand{\gammaCdf}[3]{\mathcal{GAMMA CDF}\left(#1|#2,#3\right)}
\newcommand{\gammaDist}[3]{\mathcal{G}\left(#1|#2,#3\right)}
\newcommand{\gammaSamp}[2]{\mathcal{G}\left(#1,#2\right)}
\newcommand{\gaussianDist}[3]{\mathcal{N}\left(#1|#2,#3\right)}
\newcommand{\gaussianSamp}[2]{\mathcal{N}\left(#1,#2\right)}
\newcommand{\given}{|}
\newcommand{\half}{\frac{1}{2}}
\newcommand{\heaviside}{H}
\newcommand{\hiddenMatrix}{\mathbf{ \MakeUppercase{\hiddenScalar}}}
\newcommand{\hiddenScalar}{h}
\newcommand{\hiddenVector}{\mathbf{ \hiddenScalar}}
\newcommand{\identityMatrix}{\eye}
\newcommand{\inducingInputScalar}{z}
\newcommand{\inducingInputVector}{\mathbf{ \inducingInputScalar}}
\newcommand{\inducingInputMatrix}{\mathbf{Z}}
\newcommand{\inducingScalar}{u}
\newcommand{\inducingVector}{\mathbf{ \inducingScalar}}
\newcommand{\inducingMatrix}{\mathbf{U}}
\newcommand{\inlineDiff}[2]{\text{d}#1/\text{d}#2}
\newcommand{\inputDim}{q}
\newcommand{\inputMatrix}{\mathbf{X}}
\newcommand{\inputScalar}{x}
\newcommand{\inputSpace}{\mathcal{X}}
\newcommand{\inputVals}{\inputVector}
\newcommand{\inputVector}{\mathbf{ \inputScalar}}
\newcommand{\iterNum}{k}
\newcommand{\kernel}{\kernelScalar}
\newcommand{\kernelMatrix}{\mathbf{K}}
\newcommand{\kernelScalar}{k}
\newcommand{\kernelVector}{\mathbf{ \kernelScalar}}
\newcommand{\kff}{\kernelScalar_{\mappingFunction \mappingFunction}}
\newcommand{\kfu}{\kernelVector_{\mappingFunction \inducingScalar}}
\newcommand{\kuf}{\kernelVector_{\inducingScalar \mappingFunction}}
\newcommand{\kuu}{\kernelVector_{\inducingScalar \inducingScalar}}
\newcommand{\lagrangeMultiplier}{\lambda}
\newcommand{\lagrangeMultiplierMatrix}{\boldsymbol{ \Lambda}}
\newcommand{\lagrangian}{L}
\newcommand{\laplacianFactor}{\mathbf{ \MakeUppercase{\laplacianFactorScalar}}}
\newcommand{\laplacianFactorScalar}{m}
\newcommand{\laplacianFactorVector}{\mathbf{ \laplacianFactorScalar}}
\newcommand{\laplacianMatrix}{\mathbf{L}}
\newcommand{\laplacianScalar}{\ell}
\newcommand{\laplacianVector}{\mathbf{ \ell}}
\newcommand{\latentDim}{q}
\newcommand{\latentDistanceMatrix}{\boldsymbol{ \Delta}}
\newcommand{\latentDistanceScalar}{\delta}
\newcommand{\latentDistanceVector}{\boldsymbol{ \delta}}
\newcommand{\latentForce}{f}
\newcommand{\latentFunction}{u}
\newcommand{\latentFunctionVector}{\mathbf{ \latentFunction}}
\newcommand{\latentFunctionMatrix}{\mathbf{ \MakeUppercase{\latentFunction}}}
\newcommand{\latentIndex}{j}
\newcommand{\latentScalar}{z}
\newcommand{\latentVector}{\mathbf{ \latentScalar}}
\newcommand{\latentMatrix}{\mathbf{Z}}
\newcommand{\learnRate}{\eta}
\newcommand{\lengthScale}{\ell}
\newcommand{\rbfWidth}{\ell}
\newcommand{\likelihoodBound}{\mathcal{L}}
\newcommand{\likelihoodFunction}{L}
\newcommand{\locationScalar}{\mu}
\newcommand{\locationVector}{\boldsymbol{ \locationScalar}}
\newcommand{\locationMatrix}{\mathbf{M}}
\newcommand{\variance}[1]{\text{var}\left( #1 \right)}
\newcommand{\mappingFunction}{f}
\newcommand{\mappingFunctionMatrix}{\mathbf{F}}
\newcommand{\mappingFunctionTwo}{g}
\newcommand{\mappingFunctionTwoMatrix}{\mathbf{G}}
\newcommand{\mappingFunctionTwoVector}{\mathbf{ \mappingFunctionTwo}}
\newcommand{\mappingFunctionVector}{\mathbf{ \mappingFunction}}
\newcommand{\scaleScalar}{s}
\newcommand{\mappingScalar}{w}
\newcommand{\mappingVector}{\mathbf{ \mappingScalar}}
\newcommand{\mappingMatrix}{\mathbf{W}}
\newcommand{\mappingScalarTwo}{v}
\newcommand{\mappingVectorTwo}{\mathbf{ \mappingScalarTwo}}
\newcommand{\mappingMatrixTwo}{\mathbf{V}}
\newcommand{\maxIters}{K}
\newcommand{\meanMatrix}{\mathbf{M}}
\newcommand{\meanScalar}{\mu}
\newcommand{\meanTwoMatrix}{\mathbf{M}}
\newcommand{\meanTwoScalar}{m}
\newcommand{\meanTwoVector}{\mathbf{ \meanTwoScalar}}
\newcommand{\meanVector}{\boldsymbol{ \meanScalar}}
\newcommand{\mrnaConcentration}{m}
\newcommand{\naturalFrequency}{\omega}
\newcommand{\neighborhood}[1]{\mathcal{N}\left( #1 \right)}
\newcommand{\neilurl}{http://inverseprobability.com/}
\newcommand{\noiseMatrix}{\boldsymbol{ E}}
\newcommand{\noiseScalar}{\epsilon}
\newcommand{\noiseVector}{\boldsymbol{ \epsilon}}
\newcommand{\norm}[1]{\left\Vert #1 \right\Vert}
\newcommand{\normalizedLaplacianMatrix}{\hat{\mathbf{L}}}
\newcommand{\normalizedLaplacianScalar}{\hat{\ell}}
\newcommand{\normalizedLaplacianVector}{\hat{\mathbf{ \ell}}}
\newcommand{\numActive}{m}
\newcommand{\numBasisFunc}{m}
\newcommand{\numComponents}{m}
\newcommand{\numComps}{K}
\newcommand{\numData}{n}
\newcommand{\numFeatures}{K}
\newcommand{\numHidden}{h}
\newcommand{\numInducing}{m}
\newcommand{\numLayers}{\ell}
\newcommand{\numNeighbors}{K}
\newcommand{\numSequences}{s}
\newcommand{\numSuccess}{s}
\newcommand{\numTasks}{m}
\newcommand{\numTime}{T}
\newcommand{\numTrials}{S}
\newcommand{\outputIndex}{j}
\newcommand{\paramVector}{\boldsymbol{ \theta}}
\newcommand{\parameterMatrix}{\boldsymbol{ \Theta}}
\newcommand{\parameterScalar}{\theta}
\newcommand{\parameterVector}{\boldsymbol{ \parameterScalar}}
\newcommand{\partDiff}[2]{\frac{\partial#1}{\partial#2}}
\newcommand{\precisionScalar}{j}
\newcommand{\precisionVector}{\mathbf{ \precisionScalar}}
\newcommand{\precisionMatrix}{\mathbf{J}}
\newcommand{\pseudotargetScalar}{\widetilde{y}}
\newcommand{\pseudotargetVector}{\mathbf{ \pseudotargetScalar}}
\newcommand{\pseudotargetMatrix}{\mathbf{ \widetilde{Y}}}
\newcommand{\rank}[1]{\text{rank}\left(#1\right)}
\newcommand{\rayleighDist}[2]{\mathcal{R}\left(#1|#2\right)}
\newcommand{\rayleighSamp}[1]{\mathcal{R}\left(#1\right)}
\newcommand{\responsibility}{r}
\newcommand{\rotationScalar}{r}
\newcommand{\rotationVector}{\mathbf{ \rotationScalar}}
\newcommand{\rotationMatrix}{\mathbf{R}}
\newcommand{\sampleCovScalar}{s}
\newcommand{\sampleCovVector}{\mathbf{ \sampleCovScalar}}
\newcommand{\sampleCovMatrix}{\mathbf{s}}
\newcommand{\scalarProduct}[2]{\left\langle{#1},{#2}\right\rangle}
\newcommand{\sign}[1]{\text{sign}\left(#1\right)}
\newcommand{\sigmoid}[1]{\sigma\left(#1\right)}
\newcommand{\singularvalue}{\ell}
\newcommand{\singularvalueMatrix}{\mathbf{L}}
\newcommand{\singularvalueVector}{\mathbf{l}}
\newcommand{\sorth}{\mathbf{u}}
\newcommand{\spar}{\lambda}
\newcommand{\trace}[1]{\text{tr}\left(#1\right)}
\newcommand{\BasalRate}{B}
\newcommand{\DampingCoefficient}{C}
\newcommand{\DecayRate}{D}
\newcommand{\Displacement}{X}
\newcommand{\LatentForce}{F}
\newcommand{\Mass}{M}
\newcommand{\Sensitivity}{S}
\newcommand{\basalRate}{b}
\newcommand{\dampingCoefficient}{c}
\newcommand{\mass}{m}
\newcommand{\sensitivity}{s}
\newcommand{\springScalar}{\kappa}
\newcommand{\springVector}{\boldsymbol{ \kappa}}
\newcommand{\springMatrix}{\boldsymbol{ \mathcal{K}}}
\newcommand{\tfConcentration}{p}
\newcommand{\tfDecayRate}{\delta}
\newcommand{\tfMrnaConcentration}{f}
\newcommand{\tfVector}{\mathbf{ \tfConcentration}}
\newcommand{\velocity}{v}
\newcommand{\sufficientStatsScalar}{g}
\newcommand{\sufficientStatsVector}{\mathbf{ \sufficientStatsScalar}}
\newcommand{\sufficientStatsMatrix}{\mathbf{G}}
\newcommand{\switchScalar}{s}
\newcommand{\switchVector}{\mathbf{ \switchScalar}}
\newcommand{\switchMatrix}{\mathbf{S}}
\newcommand{\tr}[1]{\text{tr}\left(#1\right)}
\newcommand{\loneNorm}[1]{\left\Vert #1 \right\Vert_1}
\newcommand{\ltwoNorm}[1]{\left\Vert #1 \right\Vert_2}
\newcommand{\onenorm}[1]{\left\vert#1\right\vert_1}
\newcommand{\twonorm}[1]{\left\Vert #1 \right\Vert}
\newcommand{\vScalar}{v}
\newcommand{\vVector}{\mathbf{v}}
\newcommand{\vMatrix}{\mathbf{V}}
\newcommand{\varianceDist}[2]{\text{var}_{#2}\left( #1 \right)}
\newcommand{\vecb}[1]{\left(#1\right):}
\newcommand{\weightScalar}{w}
\newcommand{\weightVector}{\mathbf{ \weightScalar}}
\newcommand{\weightMatrix}{\mathbf{W}}
\newcommand{\weightedAdjacencyMatrix}{\mathbf{A}}
\newcommand{\weightedAdjacencyScalar}{a}
\newcommand{\weightedAdjacencyVector}{\mathbf{ \weightedAdjacencyScalar}}
\newcommand{\onesVector}{\mathbf{1}}
\newcommand{\zerosVector}{\mathbf{0}}


\section{Introduction}\label{introduction}

There is a lot of talk about the fourth industrial revolution centered
around AI. If we are at the start of the fourth industrial we also have
the unusual honour of being the first to name our revolution before it's
occurred.

The technology that has driven the revolution in AI is machine learning.
And when it comes to capitalising on the new generation of deployed
machine learning solutions there are practical difficulties we must
address.

In 1987 the economist Robert Solow quipped ``You can see the computer
age everywehere but in the productivity statistics''. Thirty years
later, we could equally apply that quip to the era of artificial
intelligence.

From my perspective, the current era is merely the continuation of the
information revolution. A revolution that was triggered by the wide
availability of the silicon chip. But whether we are in the midst of a
new revolution, or this is just the continuation of an existing
revolution, it feels important to characterize the challenges of
deploying our innovation and consider what the solutions may be.

There is no doubt that new technologies based around machine learning
have opened opportunities to create new businesses. When home computers
were introduced there were also new opportunities in software
publishing, computer games and a magazine industry around it. The Solow
paradox arose because despite this visible activity these innovations
take time to percolate through to \emph{existing} businesses.

\subsection{Brownfield and Greenfield
Innovation}\label{brownfield-and-greenfield-innovation}

Understanding how to make the best use of new technology takes time. I
call this approach, \emph{brownfield innovation}. In the construction
industry, a brownfield site is land with pre-existing infrastructure on
it, whereas a \emph{greenfield} site is where construction can start
afresh.

The term brownfield innovation arises by analogy. Brownfield innovation
is when you are innovating in a space where there is pre-existing
infrastructure. This is the challenge of introducing artificial
intelligence to existing businesses. Greenfield innovation, is
innovating in areas where no pre-existing infrastructure exists.

One way we can make it easier to benefit from machine learning in both
greenfield and brownfield innovation is to better characterise the steps
of machine learning systems design. Just as software systems design
required new thinking, so does machine learning systems design.

In this paper we characterise the process for machine learning systems
design, convering some of the issues we face, with the 3D process:
Decomposition\footnote{In earlier versions of the Three D process I've
  referred to this as the design stage, but decomposition feels more
  appropriate for what the stage involves and that preserves the word
  design for the overall process of machine learning systems design.},
Data and Deployment.

We will consider each component in turn, although there is interplace
between components. Particularly between the task decomposition and the
data availability. We will first outline the \emph{decomposition}
challenge.

One of the most succesful machine learning approaches has been
supervised learning. So we will mainly focus on \emph{supervised
learning} because this is also, arguably, the technology that is best
understood within machine learning.

\section{Decomposition}\label{decomposition}

Machine learning is not magical pixie dust, we cannot simply automate
all decisions through data. We are constrained by our data (see below)
and the models we use. Machine learning models are relatively simple
function mappings that include characteristics such as smoothness. With
some famous exceptions, e.g.~speech and image data, inputs are
constrained in the form of vectors and the model consists of a
mathematically well behaved function. This means that some careful
thought has to be put in to the right sub-process to automate with
machine learning. This is the challenge of \emph{decomposition} of the
machine learning system.

Any repetitive task is a candidate for automation, but many of the
repetitive tasks we perform as humans are more complex than any
individual algorithm can replace. The selection of which task to
automate becomes critical and has downstream effects on our overall
system design.

\subsection{Pigeonholing}\label{pigeonholing}

The machine learning systems design process calls for separating a
complex task into decomposable separate entities. A process we can think
of as pigeonholing.

Some aspects to take into account are

\begin{enumerate}
\def\labelenumi{\arabic{enumi}.}
\tightlist
\item
  Can we refine the decision we need to a set of repetitive tasks where
  input information and output decision/value is well defined?
\item
  Can we represent each sub-task we've defined with a mathematical
  mapping?
\end{enumerate}

The representation necessary for the second aspect may involve massaging
of the problem: feature selection or adaptation. It may also involve
filtering out exception cases (perhaps through a pre-classification).

All else being equal, we'd like to keep our models simple and
interpretable. If we can convert a complex mapping to a linear mapping
through clever selection of sub-tasks and features this is a big win.

For example, Facebook have \emph{feature engineers}, individuals whose
main role is to design features they think might be useful for one of
their tasks (e.g.~newsfeed ranking, or ad matching). Facebook have a
training/testing pipeline called
\href{https://www.facebook.com/Engineering/posts/fblearner-flow-is-a-machine-learning-platform-capable-of-easily-reusing-algorith/10154077833317200/}{FBLearner}.
Facebook have predefined the sub-tasks they are interested in, and they
are tightly connected to their business model.

It is easier for Facebook to do this because their business model is
heavily focused on user interaction. A challenge for companies that have
a more diversified portfolio of activities driving their business is the
identification of the most appropriate sub-task. A potential solution to
feature and model selection is known as \emph{AutoML}
\citep{Feurer:automl15}. Or we can think of it as using Machine Learning
to assist Machine Learning. It's also called meta-learning. Learning
about learning. The input to the ML algorithm is a machine learning
task, the output is a proposed model to solve the task.

One trap that is easy to fall in is too much emphasis on the type of
model we have deployed rather than the appropriateness of the task
decomposition we have chosen.

\noindent\textbf{Recommendation}: Conditioned on task decomposition, we should automate the process of
model improvement. Model updates should not be discussed in management
meetings, they should be deployed and updated as a matter of course.
Further details below on model deployment, but model updating needs to
be considered at design time. This is the domain of AutoML.

The answer to the question which comes first, the chicken or the egg is
simple, they co-evolve \citep{Popper:conjectures63}. Similarly, when we
place components together in a complex machine learning system, they
will tend to co-evolve and compensate for one another.

To form modern decision making systems, many components are interlinked.
We decompose our complex decision making into individual tasks, but the
performance of each component is dependent on those upstream of it.

This naturally leads to co-evolution of systems, upstream errors can be
compensated by downstream corrections.

To embrace this characteristic, end-to-end training could be considered.
Why produce the best forecast by metrics when we can just produce the
best forecast for our systems? End to end training can lead to
improvements in performance, but it would also damage our systems
decomposability and its interpretability, and perhaps its adaptability.

The less human interpretable our systems are, the harder they are to
adapt to different circumstances or diagnose when there's a challenge.
The trade-off between interpretability and performance is a constant
tension which we should always retain in our minds when performing our
system design.

\section{Data}\label{data}

It is difficult to overstate the importance of data. It is half of the
equation for machine learning, but is often utterly neglected. We can
speculate that there are two reasons for this. Firstly, data cleaning is
perceived as tedious. It doesn't seem to consist of the same
intellectual challenges that are inherent in constructing complex
mathematical models and implementing them in code. Secondly, data
cleaning is highly complex, it requires a deep understanding of how
machine learning systems operate and good intuitions about the data
itself, the domain from which data is drawn (e.g.~Supply Chain) and what
downstream problems might be caused by poor data quality.

A consequence of these two reasons, data cleaning seems difficult to
formulate into a readily teachable set of principles. As a result it is
heavily neglected in courses on machine learning and data science.
Despite data being half the equation, most University courses spend
little to no time on its challenges.

\subsection{The Data Crisis}\label{the-data-crisis}

Anecdotally, talking to data modelling scientists. Most say they spend
80\% of their time acquiring and cleaning data. This is precipitating
what I refer to as the ``data crisis''. This is an analogy with
software. The ``software crisis'' was the phenomenon of inability to
deliver software solutions due to increasing complexity of
implementation. There was no single shot solution for the software
crisis, it involved better practice (scrum, test orientated development,
sprints, code review), improved programming paradigms (object
orientated, functional) and better tools (CVS, then SVN, then git).

However, these challenges aren't new, they are merely taking a different
form. From the computer's perspective software \emph{is} data. The first
wave of the data crisis was known as the \emph{software crisis}.

\subsubsection{The Software Crisis}\label{the-software-crisis}

In the late sixties early software programmers made note of the
increasing costs of software development and termed the challenges
associated with it as the
``\href{https://en.wikipedia.org/wiki/Software_crisis}{Software
Crisis}''. Edsger Dijkstra referred to the crisis in his 1972 Turing
Award winner's address \citep{Dijkstra:humble72}.

\begin{quote}
The major cause of the software crisis is that the machines have become
several orders of magnitude more powerful! To put it quite bluntly: as
long as there were no machines, programming was no problem at all; when
we had a few weak computers, programming became a mild problem, and now
we have gigantic computers, programming has become an equally gigantic
problem.

Edsger Dijkstra, The Humble Programmer
\end{quote}
We can update Dijkstra's quote for the modern era.

\begin{quote}
The major cause of \emph{the data crisis} is that machines have become
more interconnected than ever before. Data access is therefore cheap,
but data quality is often poor. What we need is cheap high quality data.
That implies that we develop processes for improving and verifying data
quality that are efficient.

There would seem to be two ways for improving efficiency. Firstly, we
should not duplicate work. Secondly, where possible we should automate
work.
\end{quote}

What I term ``The Data Crisis'' is the modern equivalent of this
problem. The quantity of modern data, and the lack of attention paid to
data as it is initially ``laid down'' and the costs of data cleaning are
bringing about a crisis in data-driven decision making. This crisis is
at the core of the challenge of \emph{technical debt} in machine
learning \citep{Sculley:debt15}.

Just as with software, the crisis is most correctly addressed by
`scaling' the manner in which we process our data. Duplication of work
occurs because the value of data cleaning is not correctly recognised in
management decision making processes. Automation of work is increasingly
possible through techniques in artificial intelligence, but this will
also require better management of the data science pipeline so that data
about data science (meta-data science) can be correctly assimilated and
processed. The Alan Turing institute has a program focussed on this
area,
\href{https://www.turing.ac.uk/research_projects/artificial-intelligence-data-analytics/}{AI
for Data Analytics}.

Data is the new software, and the data crisis is already upon us. It is
driven by the cost of cleaning data, the paucity of tools for monitoring
and maintaining our deployments, the provenance of our models (e.g.~with
respect to the data they're trained on).

Three principal changes need to occur in response. They are cultural and
infrastructural.

\subsection{The Data First Paradigm}\label{the-data-first-paradigm}

First of all, to excel in data driven decision making we need to move
from a \emph{software first} paradigm to a \emph{data first} paradigm.
That means refocusing on data as the product. Software is the
intermediary to producing the data, and its quality standards must be
maintained, but not at the expense of the data we are producing. Data
cleaning and maintenance need to be prized as highly as software
debugging and maintenance. Instead of \emph{software} as a service, we
should refocus around \emph{data} as a service. This first change is a
cultural change in which our teams think about their outputs in terms of
data. Instead of decomposing our systems around the software components,
we need to decompose them around the data generating and consuming
components.\footnote{This is related to challenges of machine learning
  and technical debt \citep{Sculley:debt15}, although we are trying to
  frame the solution here rather than the problem.} Software first is
only an intermediate step on the way to be coming \emph{data first}. It
is a necessary, but not a sufficient condition for efficient machine
learning systems design and deployment. We must move from \emph{software
orientated architecture} to a \emph{data orientated architecture}.

\subsection{Data Quality}\label{data-quality}

Secondly, we need to improve our language around data quality. We cannot
assess the costs of improving data quality unless we generate a language
around what data quality means.

\href{http://inverseprobability.com/2017/01/12/data-readiness-levels}{Data
Readiness Levels} \citep{Lawrence:drl17} are an attempt to develop a
language around data quality that can bridge the gap between technical
solutions and decision makers such as managers and project planners. The
are inspired by Technology Readiness Levels which attempt to quantify
the readiness of technologies for deployment. They contain three grades
of data readinees. Grade C data is \emph{hearsay} data. Data that is
purported to exist, but has not been electronically loaded into a
computer system (e.g.~an analysis package such as R or made available
via a REST API). Moving data from Grade C to Grade B also involves legal
and ownership issues. Once data can be loaded in it becomes Grade B.
Grade B is available electronically, but undergoes a process of
validation. Aspects like missing values, outlier representation,
duplicate records are dealt with in Grade B. Grade B also has some of
the characteristics of \emph{exploratory data analysis}
\citep{Tukey:exploratory77}. Grade A is then data in context. Finally,
at Grade A we consider the appropriateness of data to answer a
particular question. In historical statistics Grade A data might be data
that is ready for confirmatory data analysis. Many statisticians and
machine learning researchers are used to only dealing with data at Grade
A. Either because they work mainly with benchmark data or because data
was actively collected with a particular question in mind. The major
change for the era of data science is that so much data is available by
\emph{happenstance}.

\subsection{Move Beyond Software Engineering to Data
Engineering}\label{move-beyond-software-engineering-to-data-engineering}

Thirdly, we need to improve our mental model of the separation of data
science from applied science. A common trap in current thinking around
data is to see data science (and data engineering, data preparation) as
a sub-set of the software engineer's or applied scientist's skill set.
As a result we recruit and deploy the wrong type of resource. Data
preparation and question formulation is superficially similar to both
because of the need for programming skills, but the day to day problems
faced are very different.

\noindent\textbf{Recommendation}: Build a shared understanding of the language of data readiness levels
for use in planning documents, the costing of data cleaning, and the
benefits of reusing cleaned data.

\subsection{Combining Data and Systems
Design}\label{combining-data-and-systems-design}

One analogy I find helpful for understanding the depth of change we need
is the following. Imagine as an engineer, you find a USB stick on the
ground. And for some reason you \emph{know} that on that USB stick is a
particular API call that will enable you to make a significant positive
difference on a business problem. However, you also know on that USB
stick there is potentially malicious code. The most secure thing to do
would be to \emph{not} introduce this code into your production system.
But what if your manager told you to do so, how would you go about
incorporating this code base?

The answer is \emph{very} carefully. You would have to engage in a
process more akin to debugging than regular software engineering. As you
understood the code base, for your work to be reproducible, you should
be documenting it, not just what you discovered, but how you discovered
it. In the end, you typically find a single API call that is the one
that most benefits your system. But more thought has been placed into
this line of code than any line of code you have written before.

Even then, when your API code is introduced into your production system,
it needs to be deployed in an environment that monitors it. We cannot
rely on an individual's decision making to ensure the quality of all our
systems. We need to create an environment that includes quality
controls, checks and bounds, tests, all designed to ensure that
assumptions made about this foreign code base are remaining valid.

This situation is akin to what we are doing when we incorporate data in
our production systems. When we are consuming data from others, we
cannot assume that it has been produced in alignment with our goals for
our own systems. Worst case, it may have been adversarialy produced. A
further challenge is that data is dynamic. So, in effect, the code on
the USB stick is evolving over time.

Anecdotally, resolving a machine learning challenge requires 80\% of the
resource to be focused on the data and perhaps 20\% to be focused on the
model. But many companies are too keen to employ machine learning
engineers who focus on the models, not the data.

A reservoir of data has more value if the data is consumable. The data
crisis can only be addressed if we focus on outputs rather than inputs.

For a data first architecture we need to clean our data at source,
rather than individually cleaning data for each task. This involves a
shift of focus from our inputs to our outputs. We should provide data
streams that are consumable by many teams without purification.

\noindent\textbf{Recommendation}: We need to share best practice around data
deployment across our teams.  We should make best use of our processes
where applicable, but we need to develop them to become \emph{data
  first} organizations. Data needs to be cleaned at \emph{output} not
at \emph{input}.

\section{Deployment}\label{deployment}

\subsection{Continuous Deployment}\label{continuous-deployment}

Once the decomposition is understood, the data is sourced and the models
are created, the model code needs to be deployed.

To extend our USB stick analogy further, how would we deploy that code
if we thought it was likely to evolve in production? This is what
datadoes. We cannot assume that the conditions under which we trained
our model will be retained as we move forward, indeed the only constant
we have is change.

This means that when any data dependent model is deployed into
production, it requires \emph{continuous monitoring} to ensure the
assumptions of design have not been invalidated. Software changes are
qualified through testing, in particular a regression test ensures that
existing functionality is not broken by change. Since data is
continually evolving, machine learning systems require `continual
regression testing': oversight by systems that ensure their existing
functionality has not been broken as the world evolves around them. An
approach we refer to as \emph{progression testing}
\citep{Diethe:continual19}. Unfortunately, standards around ML model
deployment yet been developed. The modern world of continuous deployment
does rely on testing, but it does not recognize the continuous evolution
of the world around us.

\noindent\textbf{Recommendation}: We establish best practice around model
deployment. We need to shift our culture from standing up a software
service, to standing up a \emph{data as a service}. Data as a Service
would involve continual monitoring of our deployed models in production.
This would be regulated by `hypervisor' systems\footnote{Emulation, or
  surrogate modelling, is one very promising approach to forming such a
  hypervisor. Emulators are models we fit to other models, often
  simulations, but the could also be other machine learning modles.
  These models operate at the meta-level, not on the systems directly.
  This means they can be used to model how the sub-systems interact. As
  well as emulators we shoulc consider real time dash boards, anomaly
  detection, mutlivariate analysis, data visualization and classical
  statistical approaches for hypervision of our deployed systems.} that
understand the context in which models are deployed and recognize when
circumstance has changed and models need retraining or restructuring.

\noindent\textbf{Recommendation}: We should consider a major re-architecting of
systems around our services. In particular we should scope the use of a
\emph{streaming architecture} (such as Apache Kafka) that ensures data
persistence and enables asynchronous operation of our systems.\footnote{These
  approaches are one area of focus for my own team's reasearch. A data
  first architecture is a prerequisite for efficient deployment of
  machine learning systems.} This would enable the provision of QC
streams, and real time dash boards as well as hypervisors.

Importantly a streaming architecture implies the services we build are
\emph{stateless}, internal state is deployed on streams alongside
external state. This allows for rapid assessment of other services'
data.

\section{Conclusion}\label{conclusion}

Machine learning has risen to prominence as an approach to
\emph{scaling} our activities. For us to continue to automate in the
manner we have over the last two decades, we need to make more use of
computer-based automation. Machine learning is allowing us to automate
processes that were out of reach before.

We operate in a technologically evolving environment. Machine learning
is becoming a key component in our decision making capabilities. But the
evolving nature of data driven systems means that new approaches to
model deployment are necessary. We have characterized three parts of the
machine learning systems design process. \emph{Decomposition} of the
problem into separate tasks that are addressable with a machine learning
solution. Collection and curation of appropriate \emph{data}.
Verificaction of data quality through data readiness levels. Using
\emph{progression} testing in our \emph{deployments}. Continuously
updating models as appropriate to ensure performance and quality is
maintained.


\bibliography{../other.bib,../lawrence.bib,../zbooks.bib}

\end{document}